\def\year{2021}\relax
\documentclass[letterpaper]{article} % DO NOT CHANGE THIS
\usepackage{aaai21}  % DO NOT CHANGE THIS
\usepackage{times}  % DO NOT CHANGE THIS
\usepackage{helvet} % DO NOT CHANGE THIS
\usepackage{courier}  % DO NOT CHANGE THIS
\usepackage[hyphens]{url}  % DO NOT CHANGE THIS
\usepackage{graphicx} % DO NOT CHANGE THIS
\usepackage{natbib}  % DO NOT CHANGE THIS AND DO NOT ADD ANY OPTIONS TO IT
\usepackage{caption} % DO NOT CHANGE THIS AND DO NOT ADD ANY OPTIONS TO IT
\usepackage{microtype}
\usepackage{times}
\usepackage{latexsym}
\usepackage{graphicx}
\usepackage{amsmath}
\usepackage{booktabs}
\usepackage{algorithm}
\usepackage{algorithmic}
\graphicspath{ {Figure/} }
\usepackage{xcolor}
\usepackage{comment}
\usepackage{multirow}

\nocopyright
\title{Abusive Language Detection in Heterogeneous Contexts: \\ Dataset Collection and the Role of Supervised Attention }

\author{
Hongyu Gong,\textsuperscript{\rm 1}
Alberto Valido,\textsuperscript{\rm 2}
Katherine M. Ingram,\textsuperscript{\rm 2} \\
Giulia Fanti,\textsuperscript{\rm 3} 
Suma Bhat,\textsuperscript{\rm 1}
Dorothy L. Espelage\textsuperscript{\rm 2} \\
}
\affiliations{
\textsuperscript{\rm 1} University of Illinois at Urbana-Champaign,
\textsuperscript{\rm 2} University of North Carolina at Chapel Hill,
\textsuperscript{\rm 3} Carnegie Mellon University\\
hgong6@illinois.edu, avalia@unc.edu, ingramkm@live.unc.edu, \\
gfanti@andrew.cmu.edu, spbhat2@illinois.edu, espelage@unc.edu
\vspace{2.5em}
}

\begin{document}

\maketitle

\begin{abstract}
Abusive language is a massive problem in online social platforms. Existing abusive language detection techniques are particularly ill-suited to comments containing \textit{heterogeneous} abusive language patterns, i.e., both abusive and non-abusive parts. This is due in part to the lack of datasets that explicitly annotate heterogeneity in abusive language. We tackle this challenge by providing an annotated dataset of abusive language in over 11,000 comments from YouTube. We account for heterogeneity in this dataset by separately annotating both the comment as a whole and the individual sentences that comprise each comment. We then propose an algorithm that uses a supervised attention mechanism to detect and categorize abusive content using multi-task learning. We empirically demonstrate the challenges of using traditional techniques on heterogeneous content and the comparative gains in performance of the proposed approach over   state-of-the-art methods.
\end{abstract}

\section{Introduction}

Abusive language refers to strongly impolite and harmful language used to hurt and control a person or a group by way of harassment, insults, threats, bullying and/or trolling  \cite{waseem2017understanding}. 
Because of the nature of digital technology, abusive language can be generated anonymously and can spread to many victims in a short time, making it a serious societal concern  \cite{price2010cyberbullying}.
Due to the profound negative impact of abusive language,  many online platforms today dedicate significant resources to  its detection and categorization \cite{nobata2016abusive}. 
As the problem grows in scope and scale, so does the need for automated detection and categorization tools. 

Despite significant prior work on the automated detection of abusive language, it remains a difficult task \cite{vidgen2019challenges}. 
One important reason for this difficulty is \textit{heterogeneity} of abuse:
abusive comments often contain a combination of abusive and non-abusive language, and it can be difficult for algorithmic approaches to understand this distinction.  
Table \ref{tab:examples} illustrates examples of heterogeneous abusive comments from YouTube.
By \emph{sentence-level heterogeneity}, we mean multi-sentence comments where some sentences are abusive, and others are not.
Comment 1 shows an instance where removing the abusive sentence does not affect the meaning of the comment.
Comment 2 is more subtle; removing the underlined sentence completely changes the comment's meaning.  
By \emph{phrase-level heterogeneity}, we mean comments where only a few words are abusive, but there are no abusive full sentences.\footnote{Considering a comment with more than one independent clause, we define it  as being “phrase-level heterogeneous” if it contains at least one independent clause that is non-abusive even though the comment as a whole is labelled as abusive, or vice versa.}
Comment 4 illustrates a combination of both types of heterogeneity. 
Notably, even though the comment is abusive, it is also pro-social, in that it is defending the victim of another abusive comment. 
Hence, there are substantial subtleties associated with detecting and understanding heterogeneous abusive language---both for humans and for automated detectors.

\begin{table}[]
\captionsetup{font=normalsize}
\centering
\resizebox{0.47\textwidth}{!}{
\begin{tabular}{|l|l|l|}
\hline
\multirow{2}{*}{\begin{tabular}[c]{@{}l@{}}Sentence- \\ level \\ heterogeneity\end{tabular}} & 1 & \begin{tabular}[c]{@{}l@{}}This case really exposes Sheehy as a total fraud.  \\ \textbf{\underline{Not to mention a total man hating cunt.}} You have   \\ done a masterful (mistressful?) job of exposing \\ this lying fraudster. Love the  ending when George \\   is cited as a MGTOW Hero! He is indeed.\end{tabular}        \\ \cline{2-3} 
                                                                                         & 2 & \begin{tabular}[c]{@{}l@{}}+Username I understand. Personally, I rarely do  \\ jokes involving violence of any sort. \textbf{\underline{(Unless it}} \\ \textbf{\underline{involves Muslims!)}}   :)\end{tabular}                                                                                                                                  \\ \hline
\begin{tabular}[c]{@{}l@{}}Phrase-level \\ heterogeneity\end{tabular}                  & 3 & \begin{tabular}[c]{@{}l@{}}+Username \textbf{\underline{I always hated him}}, cause he makes \\ his opinion sound like its all facts  plus that and \\ \textbf{\underline{he sounds like a  fuckin' faggot}} \end{tabular}                                                                                                             \\ \hline
Both                                                                                     & 4 & \begin{tabular}[c]{@{}l@{}}+Username \textbf{\underline{Shut the fuck up, cunt.}} I'm a man,  \\ and I respect her a lot, because she speaks truth.   \\ She is very eloquent and a pleasure to look at. \textbf{\underline{You}} \\
\textbf{\underline{fucking idiot,}} all you do is make fun of  peoples\\  looks, did you miss the points she was making?\end{tabular} \\ \hline
\end{tabular}}
\caption{Examples of YouTube comments with heterogeneous abusive language. Abusive parts are underlined in bold text.}
\label{tab:examples}
%\vspace{-0.3in}
\end{table}

Heterogeneity does not necessarily lessen the  effects of abusive language on victims. 
Both cyber- and traditional bullies are known to sometimes engage in a combination of friendly and bullying behaviors, while still negatively affecting victims \cite{kowalski2007electronic,james2011friend}.
Similarly, workers who receive a combination of positive and negative feedback tend to experience stronger overall negative emotional reactions \cite{choi2018effects}.
Hence, it is important for automated abusive language detectors to identify abusive language couched in non-abusive language (and vice versa). 

Unfortunately, detection algorithms today struggle with heterogeneity.
The state-of-the-art approach  for automated abusive language detection relies on supervised methods that predict abusive language by learning a stacked bidirectional LSTM with attention \cite{chakrabarty2019pay}. 
However, RNNs are known to perform poorly on lengthy and/or heterogeneous data in other domains \cite{wang2018disconnected}. 
Indeed, we also find that such techniques have poor detection performance on heterogeneous comments because they struggle to identify which \emph{part} of an abusive comment makes it abusive (e.g., Fig. \ref{fig:att_example}). 

The challenges are caused in part by a dearth of datasets explicitly annotating heterogeneity of abuse. Recently Vidgen and Derczynski  \cite{vidgen2019challenges} categorize existing abusive language datasets from the natural language processing (NLP) literature, and show that the majority of existing datasets (27/50) are collected from Twitter (e.g., \cite{waseem2016hateful,davidson2017automated,golbeck2017large}),  which imposes a hard limit of 280 characters per tweet.
We hypothesize that heterogeneity may be less common in such short comments. 
Regardless, these datasets are annotated per tweet, and therefore cannot capture heterogeneity of abuse within tweets.
Although there are datasets with longer abusive comments \cite{qian2019benchmark,ribeiroevolution}, they are also typically annotated at the comment level \footnote{To the best of our knowledge, the only dataset with sentence-level annotation is \cite{de2018hate} with the group-directed abusive content limited to hate speech.
This dataset also does not include a separate annotation for the comment as a whole.}.

Because of this data availability problem, abusive language detection algorithms have been tuned (and possibly overfitted) to microblogging platforms with short comments. 
At the same time,  many  platforms, such as Facebook 
and YouTube,  attract longer comments that naturally include a broader range of abusive language patterns of interest, including heterogeneity \cite{schmidt2017survey}.\footnote{Although there exist several datasets for Facebook, they are either not in English and/or synthetic data \cite{chung2019conan}.}  

To summarize, heterogeneous abusive language detection  is difficult today because: (a) there is little labelled data in this category, and (b) even with labelled data, existing techniques are not well-suited to heterogeneous comments. In this work, we make three contributions: 

\noindent (1) We provide the first annotated dataset of over 11,000 comments in  English collected   from over 250 YouTube channels related to feminism. The annotation was performed using a theoretically-grounded abusive language taxonomy. 
    In addition to annotating each comment as abusive language or not, we also provide sentence-level annotations to study the role of heterogeneity.
    Abusive comments are categorized into types of group-directed abuse; this is a currently under-explored area owing to lack of datasets \cite{fortuna2018survey}.

    \noindent (2) Using our YouTube dataset, we demonstrate the challenges associated with using traditional abusive language techniques on heterogeneous content. 
    
    \noindent (3) We propose a model with supervised attention mechanism for detecting abusive language  and evaluate it on our dataset. This novel attention mechanism  emulates the human decision-making process in the presence of heterogeneity. Toward this, our novel attention encoder\footnote{We release the data and the code at \url{https://github.com/HongyuGong/Abusive-Language-Detection-Categorization}.
    } maps the real-valued model attention and binary human attention to the same space. 
    \noindent (4)  We show that such an explicit supervision of  attention  results in gains of over 2\% in abusive language detection ROC AUC over the best competing baseline, and similar gains on an abusive language categorization task, which aims to classify the \textit{nature} of abusive language. Additionally, although our YouTube dataset only contains annotations of sentence-level heterogeneity, we find that our approach also improves detection in instances with phrase-level heterogeneity.

\section{Related Work}
\label{sec:related}

\noindent\textbf{Abusive language detection}. Automated abusive language detectors range from supervised machine learning models built using a combination of manually crafted features such as n-grams \cite{wulczyn2017ex}, syntactic features \cite{nobata2016abusive}, and linguistic features \cite{yin2009detection,joksimovic2019automated}, to more recent neural networks \cite{park2017one,maity2018opinion}.
{The most recent studies on abuse detection have reported  state-of-the-art performance using RNNs with the attention mechanism  \cite{pavlopoulos2017deeper,chakrabarty2019pay}.} Challenges to abusive language detection include the linguistic variety and nuances \cite{nobata2016abusive,schmidt2017survey},  and the inherent biases in dataset creation \cite{vidgen2019challenges}.
%Challenges to abusive language detection include the linguistic variety and the nuances in which abusive content is expressed (e.g., abusive language may be explicit when toxic words are used, or implicit when expressions such as sarcasm are used) \cite{nobata2016abusive,schmidt2017survey}, the inherent biases in dataset creation, including the annotation quality and the degree of annotator agreement \cite{vidgen2019challenges}. This study addresses both these issues via the creation of a dataset that has heterogeneous contexts of abusive language, and reliably annotated by a team that studies abusive behaviors.

\noindent\textbf{Abusive language categorization}. There has been a lack of abusive language datasets with finer-grained taxonomies \cite{park2017one,badjatiya2017deep,fortuna2018survey}. A fine-grained abusive language classification provides insights into the nature of abusive language, permitting a more targeted mechanism for detection and intervention \cite{hoff2009cyberbullying}. 
 Common approaches to categorization  have been learning a separate classifier for each category and relying on feature engineering \cite{van2015detection,dinakar2011modeling}. In this study, we  extend prior work by creating a dataset using a taxonomy of 4 abusive categories and then using it to train a neural categorization model via multi-task learning, an approach not explored in prior work.

% A survey \cite{fortuna2018survey} shows that most available approaches to abusive language detection have relied on no taxonomy (classifying an instance to be abusive  or not, without specifying a type) or with a specified type using coarse-level taxonomies  (e.g., racism and sexism as the only two categories) \cite{park2017one,badjatiya2017deep}.  A primary reason for this is the lack of datasets with finer-grained taxonomies. A fine-grained abusive language classification provides insights into the nature of abusive language permitting a more targeted mechanism for detection and intervention \cite{hoff2009cyberbullying}. 
% Common approaches to categorization  have been learning a separate classifier for each category \cite{van2015detection,dinakar2011modeling}.
% These available studies on categorization have relied on explicit feature engineering for categorization without exploiting the power of neural models such as RNNs. In this study, we  extend prior work  by creating a dataset using a taxonomy of 4 categories of group directed abusive language and then using it to train a neural model for categorization via multi-task learning, an approach not explored in prior work.

\noindent\textbf{Attention-based models}. 
The attention mechanism is widely incorporated into neural networks to identify focus regions in inputs (e.g., when the decision hinges on the presence of key phrases). Combined with LSTMs and learned in an unsupervised manner,  attention was  found to help models achieve good performance in certain NLP applications, e.g., \cite{luong2015effective}. Recent works have also explored the use of added supervision on the attention mechanism and found it  to help machine translation with annotated alignment information  \cite{liu2016neural}, event detection with annotated arguments  \cite{liu2017exploiting} and domain transfer with human rationale \cite{bao2018deriving}.

Our approach is to use  human rationale of abuse detection towards training robust and interpretable models with supervised attention.
%Our approach is to use  human rationale of abuse detection (available as sentence-level annotations in our dataset of comments each containing multiple sentences) towards training robust and interpretable models with  supervised attention for both abusive language detection and categorization.

\section{Dataset and Annotation}
\label{sec:dataset}

Our objective is to study abusive language detection in heterogeneous settings, where  individual comments may occur as a combination of sentences with abusive and non-abusive language, illustrated via representative examples in \ref{tab:examples}.
Existing datasets are ill-suited to this task for two reasons: (1) They generally consist of comments that are too short to observe heterogeneity. 
(2) Even in longer comments that exhibit a mixture of abusive and non-abusive language, existing datasets do not include annotations that highlight the specific \textit{portion} of each comment that is abusive. 
Our goal was to build a dataset that addresses  both of these problems, through a process that was largely consistent with the recommendations of \cite{vidgen2019challenges}.\footnote{Usernames were anonymized by a generic token.}

\subsection{Data Collection}
We chose to study the YouTube platform, where comments tend to be longer.
Despite being an open platform, the  only public dataset of abusive language from YouTube is in Arabic \cite{alakrot2018dataset}.
We collected a total of $11,540$ public comments posted on $253$ YouTube channels as of May 2017,
with an average comment length of $32$ words. 
These channels were selected as the top results for the keyword \textit{feminism}. 
We used videos related to feminism  for two reasons:
(1) we observed a high occurrence of abusive language in the results, which helped us isolate the effects of class imbalance.
(2) Due to IRB restrictions, we needed a channel with limited presence of children. We used an independent service (socialbook.io), which estimated 89\% of commenters on these channels were above 17 years of age.
Our choice to focus on feminism-related videos was made purely for pragmatic reasons, and does introduce bias to the dataset.
However, one secondary benefit may be that it can aid concurrent efforts to understand the ``manosphere" \cite{ribeiroevolution}.

\subsection{Annotation Process}

Abusive language detection models trained on data annotated by experts have better performance and generalization \cite{waseem2016you};
hence, our dataset was annotated by a diverse team of  $17$ psychology students, of whom $3$ research coordinators were graduate students studying bullying and related phenomena. 
Per the recommendations of \cite{vidgen2019challenges}, the annotators and coordinators represented a range of ethnicities, genders, and mother tongues. 

We began the process with a training session, in which annotators were given our definition of abusive language, as well as examples of (non-)abusive language, including borderline cases \cite{vidgen2019challenges}.
 Because multiple definitions of abusive language are available \cite{fortuna2018survey,peter2018cyberbullying}, we chose the definition of abusive language to be ``an expression that is
intended to hurt or attack an individual or a group of people on the basis of race, appearance, gender identity, religion, or ethnicity/nationality". 
This definition was based on related literature (e.g., \cite{nobata2016abusive}) and naturally overlaps with the notions of profanity and hate speech \cite{nockleby2000hate}. 
Apart from this high-level definition of abusive language, we developed a  codebook for a taxonomy of group-directed abusive language that was theoretically informed by a synthesis paper \cite{patchin2015measuring}. 

Once the initial training was complete, we began a three-phase annotation process. 
\textit{(1) Comment-level labeling}: Each comment was first classified as either abusive or non-abusive by  
assigning it to an annotator pair, each of whom would annotate the comment independently. 
Then, the annotators discussed any disagreements in the annotation and came to consensus.
Noting the importance of context in annotation, annotators had access to the comments that preceded any given comment; this was a significant departure from previously-collected datasets \cite{vidgen2019challenges}.  In all, $27.5\%$ of comments were classified as abusive.

 \textit{(2) Sentence-level labeling}: Next, the comments were split into a total of $26,373$ sentences. Each sentence was individually labeled as either abusive or non-abusive (even if the comment had previously been labeled as non-abusive) in the context of the comment. This nested annotation provided the valuable localized human rationale for supervised attention training in our model. 
It also provided crucial insights about the nature of heterogeneity in the dataset; we found that in a multi-sentence comment, one sentence being labelled as abusive generally caused the  comment to be classified as abusive. 
In fact, there were examples of comments with abusive language being used to defend victims of other abusive language (see Table~\ref{tab:examples} for examples). Among abusive comments, $43.4\%$ are a mixture of abusive and non-abusive sentences.
Including phrase-level annotations would have increased the granularity of heterogeneity, but we focused on identifying only sentence-level heterogeneity to reduce the annotation effort of an already difficult and time-consuming process.

 \textit{(3) Comment categorization}: Abusive comments were further classified into four predefined  content categories, an aspect that is not available in a vast majority of  datasets.  Similar to \cite{founta2018large}, our annotation scheme considered four categories of group-directed abusive language: (a) gender and sexuality, (b) race, nationality and ethnicity, (c) appearance and individual characteristics, and (d) ideology, religion or political affiliation. 
 Each sentence and comment was annotated by two annotators at each stage, and the inter-annotator agreement (Cohen's $\kappa$) was $88\%$ for the abusive sentence labeling task, $90\%$ for the abusive comment labelling, and $93\%$ for the categorization task. Annotators later met to resolve their differences. A total of $1979$ abusive comments were labelled into these categories.

\noindent\textbf{Group meetings} 
We led weekly annotation team meetings to discuss disagreements in annotations between each pair of annotators. 
The team would come to consensus as a group by consulting third-party resources (e.g., Urban Dictionary) and by relying on the diverse cultural contexts of the group.
These meetings had several benefits: \\
(1) Annotator subjectivity is known to be a major factor affecting the quality of labels for machine learning pipelines \cite{hube2019understanding}. 
Discussing difficult-to-classify comments, particularly among a culturally-diverse group of annotators, was therefore important for ensuring the quality of our annotations \cite{vidgen2019challenges}. \\
\noindent (2) Miceli \textit{et al.} recently demonstrated that annotators tend to view the opinions of their supervisors as authoritative, and defer to their judgment  \cite{miceli2020between}. 
We explicitly aimed to avoid such an effect by having annotators guide \textit{each other} to consensus. \\
(3) Our weekly meetings were also designed to monitor the emotional health of our annotators. 
Abusive language is known  to have negative psychological effects on bystanders as well as victims \cite{low2007experiences,ferguson2011know}. 
We observed such effects among our research assistants, who described the experience as ``a taxing process, psychologically" and the content as ``appalling." 
We therefore discussed the annotators' emotional state at each meeting.

\begin{figure}[h!]
\captionsetup{font=normalsize}
\centering
\includegraphics[width=0.4\textwidth]{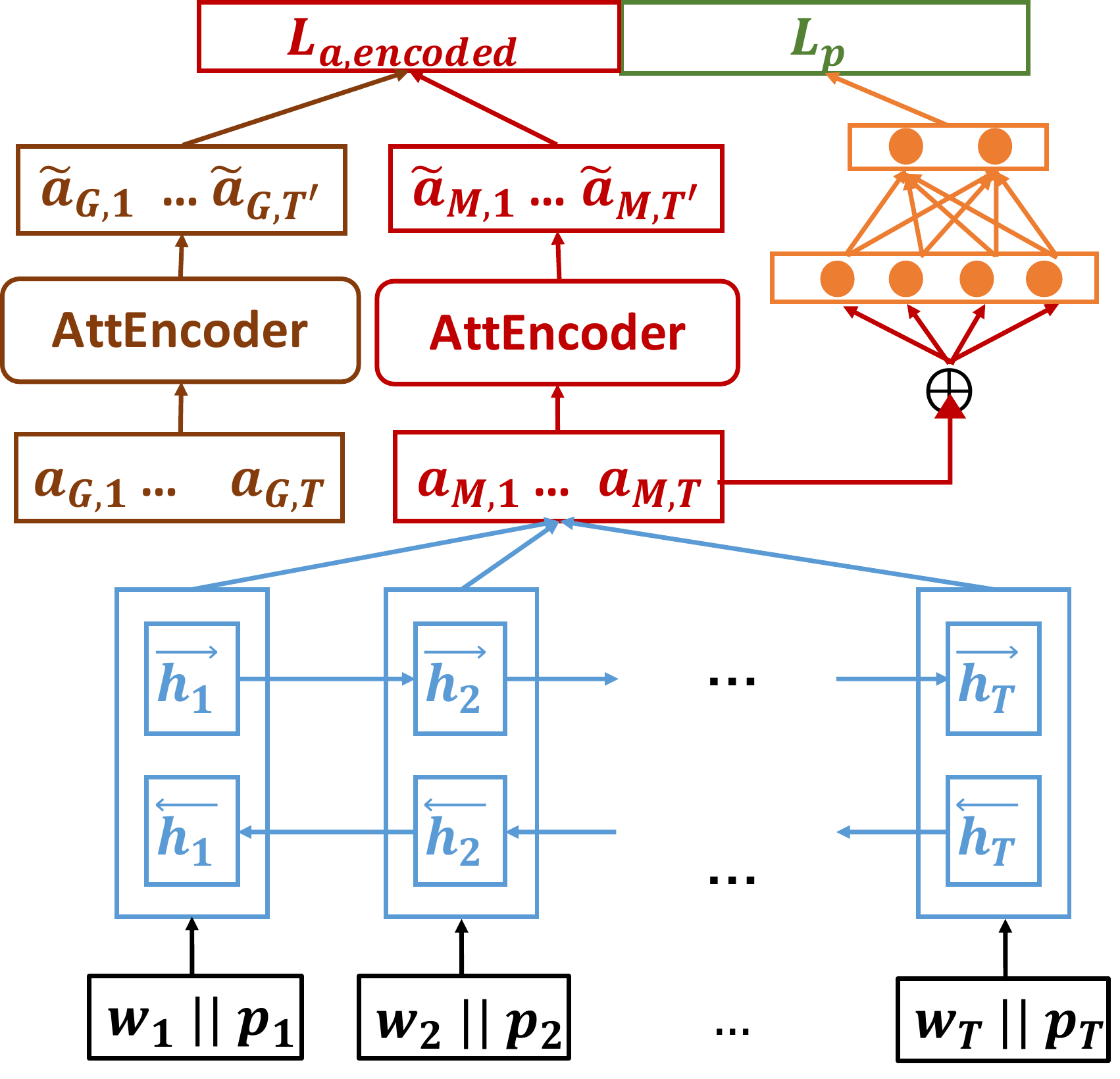}
\caption{RNN with supervised attention. Each input word is a concatenation of its word- and POS embeddings. The input is a sequence of embeddings, and the recurrent layer generates hidden state vectors. The attention module determines the attention distribution over these hidden vectors, which are linearly weighted with the attention weights to  input to the feedforward network (FFN). The FFN  predicts the scores of each class, which are transformed to a probability distribution over the classes in the output. The AttEncoder maps the ground truth (G) and model (M) attention to the same vector space in order to measure the encoded attention loss.}
\label{rnn_supervised_attetnion}
%\vspace{-6mm}
\end{figure}

\section{Abusive Language Detection}
\label{sec:detection}
About $43.4\%$ of the abusive comments in our YouTube data are a mixture of both abusive and non-abusive content. We hypothesize that manually demarcating the abusive portion in a comment from other non-abusive content (as a way of showing the human rationale) can provide better supervision while training  abusive language detectors.

\begin{figure*}[htbp!]
\captionsetup{font=normalsize}
\centering
\begin{minipage}[c]{0.65\textwidth}
\centerline{\includegraphics[width=\linewidth]{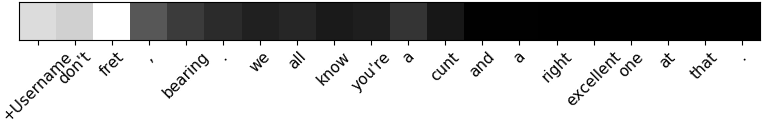}}
\centerline{\small{(a) No supervision.}}
\end{minipage}
\begin{minipage}{0.65\textwidth}
\centerline{\includegraphics[width=\linewidth]{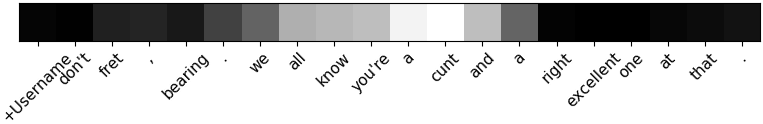}}
\centerline{\small{(b) Encoded supervision.}}
\end{minipage}
%\vspace{-2mm}
\caption{Attention visualization of RNN trained with and without attention supervision for an abusive sentence in grey-scale image. The lighter the word, the higher its attention weight.}
\label{fig:att_example}
%\vspace{-5mm}
\end{figure*}

Attention-based models available in existing literature train the attention component implicitly since only the predictive function (and not the attention mechanism) is supervised. 
Instead, we conjecture that by explicitly supervising the attention mechanism we might be able to steer the model towards learning the relevant patterns based on human rationale. For example, given a comment ``+Username don't fret, bearing, all know you're a cunt and a right excellent one at that'', we visualize the (implicitly trained) attention weights of a Recurrent Neural Network (RNN) over all the words as shown in Fig.~\ref{fig:att_example}(a). 
We see that the model classified the comment as abusive because it wrongly considered \textit{fret} as a signal of abuse instead of \textit{cunt}.

Our goal in this study is to introduce the idea of supervised attention  and train the neural network not only to give correct predictions but also to accurately identify abusive patterns from the input. We will show next that supervised attention is beneficial to both abusive language detection and categorization.

\subsection{RNN with Supervised Attention}
In this study, we use a Recurrent Neural Network with bidirectional LSTM units (a BiLSTM network) owing to its state-of-the-art performance in abusive language classification \cite{chakrabarty2019pay}. 
Let  $\{w_{1}, w_{2}, \ldots, w_{T}\}$ be the  $T$ words of a comment. 
%and accordingly their have embeddings $\{v_{1}, v_{2}, \ldots, v_{T}\}$.
Inspired by factored neural network models which incorporate extra-linguistic information, we use the part-of-speech (POS) tags of the words  $\{p_{1}, p_{2}, \ldots, p_{T}\}$ in addition to the words as the input sentence \cite{sennrich2016linguistic}.
For every input word $w_{t}$, we concatenate its word embedding and POS embedding as its vector representation $\mathbf{x}_{t}$. The word and POS embeddings are pretrained with the word2vec CBOW model on the training data \cite{mikolov2013distributed}.
The structure of our RNN model with attention supervision is shown in Fig.~\ref{rnn_supervised_attetnion}.

The BiLSTM recurrent layer incorporates contextual information from both sides of a given word; it concatenates the two state vectors $\overrightarrow{\mathbf{h}}_{t}$ and $\overleftarrow{\mathbf{h}}_{t}$ to yield the bidirectional vector $\mathbf{h}^{b}_{t}$, i.e., $\mathbf{h}^{b}_{t}=[\overrightarrow{\mathbf{h}}_{t},\overleftarrow{\mathbf{h}}_{t}]$.
We apply the attention mechanism proposed in \cite{yang2016hierarchical} and also used in \cite{chakrabarty2019pay}. The model's attention value $a_{M,t}$ on the $t$-th token is calculated from hidden state vector $\mathbf{h}^{b}_{t}$:
\begin{align}
\mathbf{v}_{t} &= \sigma(\mathbf{W}_{u}\mathbf{h}^{b}_{t} + \mathbf{b}_{u}), \\
a_{M,t} &= \frac{\text{exp}(\mathbf{u}^{T}\mathbf{v}_{t})}{\sum\limits_{t'}\text{exp}(\mathbf{u}^{T}\mathbf{v}_{t'})},
\end{align}
where $\mathbf{W}_{u}$ is a trainable matrix, $\mathbf{b}_{u}$ and $\mathbf{u}$ are vectors in the recurrent layer, and $\sigma(\cdot)$ is the sigmoid function. 
The hidden state vectors at different time steps are linearly weighted to yield the vector $\mathbf{z}=\sum\limits_{t}a_{M,t}\mathbf{h}_{t}^{b}$, a compressed representation of the input sentence. We have two FFN layers with the sigmoid activation and an output layer above the recurrent layer, which  outputs  $\mathbf{y}'$ as the model's predicted probability distribution over the classes.

The cross-entropy prediction loss $L_{p}$ is used to measure the difference between the predicted vector $\mathbf{y}'$ and the ground truth vector $\mathbf{y}$, where $\mathbf{y}'$ and $\mathbf{y}$ are two-dimensional vectors for the binary abusive classification task.

To supervise the attention over the input sentences and evaluate how well the neural network can target the truly abusive segments, we define the attention loss $L_{a}$ as part of our training objective. To indicate the abusive part in a comment, we assign a (ground-truth) binary attention vector $\bf{a}_{G}$ to the input sequence, where the words in the abusive segment are marked as 1 and others as 0. This attention vector is derived from the sentence-level labels in our data (see the section of Dataset and Annotation) and is used to train the attention mechanism. The ground-truth attention $\bf{a}_{G}$ is scaled by $1/(\bf{1}^{T}\bf{a}_{G})$ to normalize the weights. Without loss of generality, we continue to use $\bf{a}_{G}$ to refer to the scaled ground truth attention.  We use $\bf{a}_{M}$ to refer to  the attention vector output from the attention module of our system, that is   based on the outputs of the RNN. 

Ideally the estimated attention $\bf{a}_{M}$ should correspond to the ground truth attention $\bf{a}_{G}$, i.e., higher model attention weights should be assigned to the words that are marked with non-zero values. Toward this goal of aligning $\bf{a}_{M}$ and  $\bf{a}_{G}$, we consider two simple but commonly used losses in neural network training: the L1 loss ($L_{a,l_1}$), and the L2 loss ($L_{a,l_2}$). The use of the L2 attention loss has been  explored in several previous works, including \cite{liu2016neural,liu2017exploiting}.

\begin{table*}[htbp!]
\captionsetup{font=normalsize}
\centering
\resizebox{0.85\textwidth}{!}{
\begin{tabular}{|c|c|c|c|c|c|c|c|c|c|}
\hline
 & \multicolumn{4}{c|}{SVM} & \multicolumn{2}{c|}{RNN baseline with attention} & \multicolumn{3}{c|}{RNN with attention supervision (C+S)} \\ \hline
Train data & \multicolumn{2}{c|}{C} & \multicolumn{2}{c|}{C+S} & {\qquad C \qquad} & C+S & Encoded loss & L1 loss & L2 loss \\ \hline
Sentiment & Yes & No & Yes & No & No & No & No & No & No \\ \hline
ROC AUC & 0.756 & 0.750 & 0.782 & 0.774 & 0.796 & 0.803 & \textbf{0.826} & 0.814 & 0.810 \\ \hline
PR AUC & 0.581 & 0.572 & 0.614 & 0.606 & 0.585 & 0.633 & \textbf{0.654} & 0.638 & 0.636 \\ \hline
F1 score & 0.554 & 0.545 & 0.544 & 0.540 & 0.584 & 0.615 & \textbf{0.624} & 0.618 & 0.608 \\ \hline
\end{tabular}}
\caption{Abusive language detection performance. C:  using only comment labels, and C+S:  using comment and sentence labels. }
\label{tab:detection_performance}
%\vspace{-0.6cm}
\end{table*}

\noindent\textbf{Encoding attention.} 
Because our ground truth annotation  does not mark the degree to which words are abusive, all the words in an abusive segment are assigned non-zero weights. More specifically, the components of the ground truth attention vector are either zero or uniformly non-zero. 
In contrast, the model attention is real valued, assigning different weights to the words.
As a result, we note that the attention vectors  $\bf{a}_{M}$ and $\bf{a}_{G}$ are not in the same vector space, and hence computing the L1 and L2 losses over these attention vectors as is may not accurately capture their correspondence. Toward remedying this situation, we propose an attention encoder (termed as \textit{AttEncoder}) to encode the ground truth and the model attention vectors to a common space and to then estimate how well they match. This is done via a neural module, which is shown in Fig.~\ref{rnn_supervised_attetnion}. 
The ground truth attention $\bf{a}_{G}$ is encoded as $\tilde{\bf{a}_{G}}$ using an AttEncoder as indicated below.
\begin{align}
\tilde{\mathbf{a}}_{G} &= \text{AttEncoder}_{G}(\mathbf{a}_{G}) 
= \text{tanh}(\mathbf{W}_{G}\mathbf{a}_{G} + \mathbf{b}_{G}), 
\label{eq:proj_ground_attention}
\end{align}
where $\text{tanh}(\cdot)$ is an activation function, matrix $\mathbf{W}_{G}$ and bias vector $\mathbf{b}_{G}$ are tunable parameters.

Similarly, the model attention $\mathbf{a}_{M}$ is transformed by another AttEncoder:
\begin{align}
\label{eq:proj_model_attention}
\tilde{\mathbf{a}}_{M} &= \text{AttEncoder}_{M}(\mathbf{a}_{M}) 
= \text{tanh}(\mathbf{W}_{M}\mathbf{a}_{M} + \mathbf{b}_{M}),
\end{align}
where $\mathbf{W}_{M}$ and $\mathbf{b}_{M}$ are tunable parameters. 
The resulting attention vectors $\tilde{\mathbf{a}}_{G}$ and $\tilde{\mathbf{a}}_{M}$ are in the same hidden space and their inner product can be interpreted as their similarity. The encoded attention loss estimated by this  module is $L_{a,\text{encoded}}$:
\begin{align}
L_{a,\text{encoded}} = -\tilde{\mathbf{a}}_{G}^{T}\tilde{\mathbf{a}}_{M}.
\end{align}

The total loss $L$ is defined to be a weighted sum of the prediction loss $L_p$ and the attention loss $L_a$: 
\begin{align}
L = L_{p} + \beta L_{a},
\end{align}
{where $L_{a}$ can be one of the three attention losses, $L_{a,1}$, $L_{a,2}$ and $L_{a,\text{encoded}}$.}
The hyperparameter $\beta$ is tuned on the validation set, and $\beta=0.2$ in our experiments. The neural network is trained  from end to end to minimize the total loss with explicit attention supervision $L_{a}$, thereby capturing abusive patterns and making classification decisions.

\subsection{Experiments on Abuse Detection}

We compare our system with previous models used for abusive language detection. 
In our experiments, we randomly split the annotated data (at the comment-level) into training, validation and test sets in a ratio of 3:1:1 for use in the  abuse detection and categorization tasks.
The baselines are: \\
\noindent(1) Support vector machine (SVM)---included in this comparison because of its competitive performance in related prior works \cite{van2015detection,nobata2016abusive}. It takes word unigrams, bigrams and character trigrams as features. The vocabulary sizes of word- and character- ngrams are 5,000.
We also use the sentiment feature from \cite{van2015detection}, where we count the number of positive, negative, and neutral words in the input as well as the average of the lexical polarity as four numeric \textit{sentiment features}, using an opinion lexicon  provided by the NLTK package  \cite{opinionlexicon}. \\
\noindent(2) RNN with attention achieves state-of-the-art results on abuse detection on multiple public datasets as reported in \cite{chakrabarty2019pay}. Although structurally it is similar to our model in Fig.~\ref{rnn_supervised_attetnion}, it lacks the novel attention encoder and the  attention supervision that we propose in our model.

\noindent\textbf{Preprocessing}. 
We normalized the YouTube comments with a text normalization tool \cite{tweet_preprocessor}, replaced  emoticons and urls  with special symbols, and  finally tokenized and (POS) tagged the comments with the CMU social text tagging tool \cite{owoputi2013improved}.

\noindent\textbf{Evaluation metrics}. 
We consider the label ``abusive'' to be the positive class and use three metrics in our evaluation---the area under the receiver operating characteristic curve (ROC AUC), the area of the precision-recall curve (PR AUC) and the F1 score. 
The ROC AUC measures the area under the true positive vs. false positive rate curve. The PR AUC measures the area under precision vs. recall curve.
PR AUC is known to be a better metric than ROC AUC at comparing algorithms when  negative samples (benign comments in our case) are much more than positive samples (abusive comments) \cite{davis2006relationship}. Unlike F1 score that requires a specific decision-making threshold set on the test data,  ROC AUC and PR AUC are free from any threshold tuning.

\subsection{Detection Results} We train and evaluate all systems on $5$ train-test splits to reduce randomness, and report
the average performance  in Table~\ref{tab:detection_performance}. Since we have both comment-level and sentence-level annotations, {we report the performance of SVM and RNN baselines trained on labeled comments alone (denoted as ``C'' in Table~\ref{tab:detection_performance}) and the performance by using both labeled comments and labeled sentences (denoted as ``C+S'').} As for the proposed RNN with attention supervision, it is only trained on the labeled comments with access to the labels of the components sentences.

We note that RNNs always outperform the SVM classifier.
For the SVM classifier, we find that adding sentiment information does not lead to obvious improvements. By comparing the models trained on C alone, and on C+S, we observe that sentence-level annotations improve P-R AUC for SVM and the RNN baseline.

We observe that  attention supervision  makes better use of sentence-level annotations given that the RNN with encoded attention loss outperforms the RNN baseline trained on C+S by $2.3\%$ in ROC AUC, by $2.1\%$ in PR AUC and $0.9\%$ in F1 score. The performance gains are statistically significant at p-value of $0.05$ using Student's t-test.  Moreover, for the model with supervised attention, it is notable that the use of encoded loss is better than both L1 and L2 loss. The gains of the model using encoded attention loss over model instances trained with L1 or L2 are also statistically significant. 

\subsection{Attention Evaluation}
We saw how the model trained with attention supervision resulted in improved abuse detection. 
To provide a comprehensive view of the model's performance,   we evaluate the model's  ability to learn the correct abusive patterns, which is reflected in segments with high attention.

\noindent\textbf{Qualitative evaluation}. 
We evaluate models' attention on sentence segments.
The attention assigned by the RNN model \textit{without} attention supervision is depicted in Fig.~\ref{fig:att_example}(a). We compare this with the attention distribution of the model trained with encoded attention loss. As shown in Fig.~\ref{fig:att_example}(b), the abusive pattern ``you're a cunt'' was captured by the model with encoded attention loss. Notably, this example illustrates how  our annotations at the sentence-level, also help with phrase-level heterogeneity as well. 
\begin{figure}[htbp!]
\captionsetup{font=normalsize}
\centering
\includegraphics[width=0.45\textwidth]{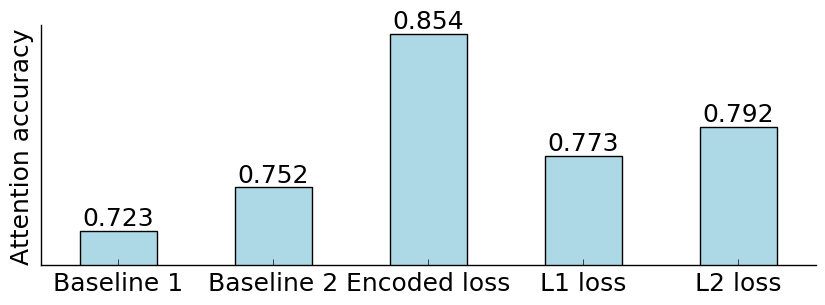}
%\vspace{-1mm}
\caption{Attention evaluation on test comments.}
\label{fig:attn_eval_plot}
%\vspace{-2mm}
\end{figure}

\noindent\textbf{Quantitative evaluation}. Next we quantitatively evaluate the predicted attention over the test comments by analyzing the model's attention weights over the component sentences of the comments. We average the attention weights of the words within each sentence to yield the sentence attention weight. For each abusive comment, we select the sentence with the highest attention weight as the predicted abusive sentence. Then we evaluate the accuracy of the abusive sentence prediction by comparing with the gold labels, yielding the percentage of automatically selected sentences which were manually annotated as abusive.

\begin{table*}[htbp!]
\captionsetup{font=normalsize}
\centering
\resizebox{0.9\textwidth}{!}{
\begin{tabular}{|c|c|c|c|c|c||c|c|c|c|c|}
\hline
 & \multicolumn{5}{c||}{Multi-task} & \multicolumn{5}{c|}{Single-task} \\ \hline
Attention supervision & Encoded & L1 & L2 & {\ Baseline 1} & {\ Baseline 2} & Encoded & L1 & L2 & {\ Baseline 1} & {\ Baseline 2}\\ \hline
Gender & \textbf{0.643} & 0.601 & 0.613 & 0.585 & 0.608  & 0.609 & 0.599 & 0.601 & 0.576 & 0.582 \\ \hline
Race & \textbf{0.551} & 0.505 & 0.503 & 0.483 &  0.505 & 0.354 & 0.323 & 0.330 & 0.168 & 0.307 \\ \hline
Appearance & \textbf{0.788} & 0.760 & 0.773 & 0.752 & 0.760 & 0.755 & 0.745 & 0.737 & 0.738 & 0.733 \\ \hline
Ideology & \textbf{0.610} & 0.577 & 0.559 & 0.496 & 0.508 & 0.511 & 0.524 & 0.512 & 0.477 & 0.499 \\ \hline
\end{tabular}}
\caption{PR AUC of abuse categorization with and without attention supervision in single- and multi-task settings.}
\label{tab:categorization}
%\vspace{-0.3cm}
\end{table*}

We report the accuracy of the models with encoded, L1 and L2 loss in Fig.~\ref{fig:attn_eval_plot}, including the two baselines trained without attention supervision---baselines 1 and 2 as   the RNN with attention, trained using C and C+S respectively.

We note that  baseline 2 captures the abusive patterns more accurately than baseline 1, showing that sentence-level annotation helps abusive segment detection. It is also noteworthy that the model with encoded attention loss outperforms baseline 2 (trained without attention supervision). Even though baseline 2 used both comment- and sentence-level labels, it was trained on isolated sentences without considering the contextual information. This highlights the effectiveness of attention supervision for learning the abusive patterns  in the context of the entire comment.

\section{Abusive Language Categorization}
\label{sec:categorization}

A fine-grained categorization of abusive comments provides insights into the nature of  abusive language.
We  manually classified the abusive comments into the category set $C=\{\text{gender}, \text{race}, \text{appearance}, \text{ideology}\}$.

\subsection{Model}
Previous work on categorization trained a classifier for each category independently \cite{van2015detection,dinakar2011modeling}. 
However, poorly represented categories (e.g., \textit{race} in our data) make  training a good classifier for such a category  difficult. We adopt the technique of multitask learning, where the main idea is to share information among multiple related tasks so as to improve the model's  generalizability  of the individual tasks \cite{standley2020tasks}. In our multitask model, the different categories share information by sharing their lower-level layers (i.e. embeddings in the input layer and the recurrent layer).
The predictions for each category are made separately in their respective output layers. We empirically show the resulting performance gain for all categories. Notably, we find that supervised attention helps not only in abuse detection but also in categorization.

The model for categorization was similar to that used for abuse detection (Fig.~\ref{rnn_supervised_attetnion}), except  the single two-dimensional output vector $\bf{y}'$ was  replaced with four two-dimensional vectors $\{\bf{y}_{c}'\}_{c\in C}$,  each two-dimensional vector $\bf{y}_{c}'$ corresponding to category $c$. We used cross-entropy loss as category $c$'s prediction loss $L_{c}$. The total loss  was again the sum of the prediction loss and the attention loss:
\begin{align}
L = \sum_{c\in C}\omega_{c}L_{c} + \beta L_{a},
\end{align}
where $\omega_{c}$ is the weight of category $c$, and $\sum\omega_{c}=1$. 
In multitasking, there is a primary category $c$ with a higher weight $\omega_{c}$ than the weights $\omega_{c'}$ for the  auxiliary categories $c'$. 
We report the per-category performance by taking each category as the primary category respectively.
The hyperparameters were tuned on the validation data, with $\beta=0.2$, $\omega_{c}=0.7$, and $\omega_{c'}=0.1, \forall c'\not=c$.

\subsection{Experiments}

As before, for our experiments on \textit{categorizing} abusive language,  we used a standard RNN model with attention as a strong baseline. Baseline 1 was trained on C, and baseline 2 was trained on C+S.
A third model is an RNN model with the same idea of attention supervision (used for the classification task) but now in a multitask learning set-up described above.

We evaluated the models with $5$  train-test splits, and report their average performance in Table~\ref{tab:categorization}. All the systems were RNNs with different attention losses in either a single-task or a multi-task setting. We report the PR AUC of each category for each system, and evaluate how supervised attention and multitask learning affect the  performance. Overall, baseline 2 achieves better PR AUC than baseline 1 due to the extra sentence-level annotations. Attention supervision with encoded loss makes better use of sentence annotations than systems with other attention losses as well as the baselines without attention loss.

Comparing the models with and without attention supervision, we note that attention supervision improves categorization in both single- and multi-tasking scenarios (all are absolute gains);  the highest improvement was seen in the poorly represented categories of \textit{race} and \textit{ideology}. For the \textit{race} category, the supervision with encoded loss improves the PR AUC  by $4.7\%$ over baseline 2 in single tasking, and $4.6\%$ in multitasking. As for \textit{ideology}, the encoded attention loss yields a gain of $10.2\%$ over baseline 2 in multitasking.

Multi-task learning improves categorization in all categories; we see an increase of $19.7\%$  in the performance of the race category when encoded attention loss is applied, an increase of $31.5\%$  in baseline 1, and an increase of $19.8\%$ in baseline 2. Note that all gains reported are absolute.

The best-performing system is the combination of encoded attention loss with multi-task learning. It uses essentially the same training data as baseline 2. Compared with baseline 2 without attention supervision in single tasking,  it increases the PR AUC by $6.1\%$ in the gender category, $24.4\%$ in race,  $5.5\%$ in appearance, and  $11.1\%$ in ideology.

\section{Conclusion and Limitations}
\label{sec:conclusion}
We have presented a new annotated dataset of abusive language from YouTube, as well as an  empirical study on the use of supervised attention of neural networks to improve the detection and categorization of abusive language.
%as well as an  empirical study on the use of supervised attention  of  neural  networks for detecting and categorizing abusive language. We show how  a model's attention can be informed by human rationale  to improve detection and base the decision on meaningful abusive patterns. 

A primary limitation of our methodology is that our data comes only from feminism-related channels, which introduced bias and limits the generality of our results. 
Moreover, due to limitations of the annotation interface, the thread structure was not available to annotators, and they did not follow links in the comments or view the associated videos.  
This was intentional, so that the automatic detection would be based solely on textual information. 
Hence, two important directions for future work are to (a) study the performance of supervised attention on a broader class of datasets, and (b) conduct a joint analysis of  text \textit{and} the accompanying media.

\section{Acknowledgements}
This work was supported in part by the National Science Foundation under grant no. 1720268.
We would like to thank our annotators: Cagil Torgal,  Ally Montesino, Lauren Fisher, Kaylie Skinner, Lital Hartzy, Madison Kohler, Gabriele Mamone, Abigail Matterson, Hannah Phillips, Palmer Tirrell, Victoria Wiliams, Kristina Youngson, Huibin Zhang and Talia Akerman, and our participants: Luz Robinson, America El Sheikh, Uma Kumar, Savannah Herrington, Briana de Cola, Ciara Tobin, Angela Rodriguez, Carmen Florez, Sky Martin, Caroline Spitz, Claudia Rodriguez and Paige Hespe.
We would also like to thank Sreedhar Radhakrishnan and Ganesh Ramadurai for their help in ensuring the reproducibility of our results and comparison against data from Stormfront.

\bibliography{aaai.bib}

\end{document}